\documentclass{article}

\usepackage{arxiv}

\usepackage[utf8]{inputenc} % allow utf-8 input
\usepackage[T1]{fontenc}    % use 8-bit T1 fonts
\usepackage{hyperref}       % hyperlinks
\usepackage{url}            % simple URL typesetting
\usepackage{booktabs}       % professional-quality tables
\usepackage{amsfonts}       % blackboard math symbols
\usepackage{nicefrac}       % compact symbols for 1/2, etc.
\usepackage{microtype}      % microtypography
\usepackage{tikz}
\usepackage{lipsum}
\usepackage{graphicx}
\usepackage{listings}
\usepackage{pgfplots}
\usepackage{float}
\pgfplotsset{compat=1.18}

\graphicspath{ {./images/} }

\title{An Exploration of Self-Supervised Mutual Information Alignment for Multi-Task Settings}

\author{
 Soham Govande \\
  Stanford University\\
  \texttt{govande@stanford.edu} \\
}

\begin{document}
\maketitle
\begin{abstract}
There is a growing need for pluralistic alignment methods that can steer language models towards individual attributes and preferences. One such method, Self-Supervised Alignment with Mutual Information (SAMI), uses conditional mutual information to encourage the connection between behavioral preferences and model responses. We conduct two experiments exploring SAMI in multi-task settings. First, we compare SAMI to Direct Preference Optimization (DPO) on a multi-task benchmark (MT-Bench), using a stronger model to generate training data for a weaker one across diverse categories (humanities, STEM, extraction, coding, math, reasoning, and roleplay). Our results indicate that one iteration of SAMI has a 57\% win rate against DPO, with significant variation in performance between task categories. Second, we examine SAMI's impact on mathematical accuracy (GSM-8K) relative to supervised fine-tuning (SFT). While SAMI increases zero-shot performance by 1.1\%, SFT is more effective with a 3.2\% boost. However, SAMI shows interesting scaling trends. When given 10 attempts, SAMI improves accuracy by 3.9\%, while SFT achieves a 10.1\% increase. Combining SAMI with SFT yields an additional improvement of 1.3\% in multi-attempt settings, though single-attempt accuracy remains unchanged.

\end{abstract}

\section{Introduction}
Pluralistic steerability is important for models to be able to adapt their responses to diverse user preferences and contexts. An exciting question is whether base language models already inherently possess this capability. Previous work has demonstrated that Self-Supervised Alignment with Mutual Information (SAMI) can enhance this steerability on tasks such as summarization and dialogue  \cite{franken_self-supervised_2024}. For instance, iterative fine-tuning on a base model’s own generations can allow models like \texttt{llama3-70b} to adopt specific styles, such as speaking like a pirate and summarizing more concisely. In this work, we further explore the SAMI algorithm, specifically investigating two questions: (1) Can SAMI be used to fine-tune a weaker model using generations from a stronger aligned model, particularly in multi-task settings like MT-Bench? and (2) Can SAMI improve chain-of-thought reasoning for mathematical accuracy? For both of these experiments, we also publish our \href{https://github.com/sohamgovande/sami-extension}{code repository} as a technical worklog.

\section {Experiments}

\subsection{Experiment 1: MT-Bench}
\textbf{Overview. }We attempt to improve \texttt{llama3.1-8b}’s performance on diverse tasks through the SAMI algorithm, and compare the performance of SAMI versus Direct Preference Optimization (DPO). We curate a collection of diverse datasets and evaluate on MT-Bench, which is a diverse, multi-task benchmark modeled after real-world user queries, covering eight categories (humanities, STEM, extraction, coding, math, reasoning, and roleplay) \cite{zheng_judging_2023}. Our goal is to determine how the mutual information alignment done by SAMI enables a model affects task performance, relative to DPO.

\textbf{Principle Generation. }A constitution consists of a set of 2 principles, with each principle being a positive instruction (typically desirable) or an antithesis (typically undesirable). For example, for mathematics, the constitution may consist of a first principle, either “write responses in LaTeX” or  “write responses in plaintext”, and a second principle, either “think step by step”, or “emit a concise final answer.” The principles for each category are described in Appendix B.1.

\textbf{Data Generation. }To improve along a diverse set of tasks, we curate an ensemble of datasets spanning the eight MT-Bench categories: \verb|lighteval/MATH|, \verb|AlekseyKorshuk/roleplay-io|, \verb|perlthoughts/coding-prompts-small|, \verb|openai/summarize_from_feedback|, \verb|livebench/reasoning|, \verb|STEM-AI-mtl/Electrical-engineering|, \verb|reasoning-machines/gsm-hard|, and \verb|allenai/WildChat-nontoxic|. We sample up to 3,000 prompts from each dataset and combine them into a final dataset of 19,663 prompts. In this final dataset, for each prompt, we randomly sample a relevant constitution consisting of two principles depending on the question’s category. We generate completions using the prompt described in Appendix B.1
. 

We initially attempted self-generating the data with the model being trained (i.e., \verb|Llama-3.1-8B-Base|). However, we empirically found that a weak model does provide enough signal to bootstrap from. Hence, we opted to generate data with a stronger model (i.e., \verb|Llama-3.1-8B-Instruct|), and distill this information into a weaker model.

% \makebox[\textwidth]{%
% \begin{tikzpicture}[every node/.style={outer sep=2}]
%     % Nodes
%     \node (base) at (-2.5, 0) {Llama3.1-8B Base};
%     \node (sft) at (0, 0) {SFT};
%     \node (sami) at (1, 1) {SAMI};
%     \node (dpo) at (1, -1) {DPO};

%     % Arrows
%     \draw[->] (base) -- (sft);
%     \draw[->] (sft) -- (sami);
%     \draw[->] (sft) -- (dpo);
% \end{tikzpicture}
% }

\textbf{Training. } Prior to performing DPO, we perform SFT using positive samples from the dataset to prevent the base model from being out of distribution \cite{rafailov_r_2024}. From this SFT baseline, we create two model variants: one trained via DPO for one epoch, and one trained via one iteration of SAMI. All training hyperparameters and Weights \& Biases runs available in Appendix C. At the end, we have three models to compare: a SFT model, a model trained via DPO, and a model trained via SAMI.

\textbf{Evaluation. }The goal of evaluation was to determine if the following relationship held true: SAMI performance > DPO performance > SFT performance. Performance was measured by a LLM judge (GPT-4) in two axes: principle alignment and principle-free judgment. Principle alignment aimed to measure whether a model’s outputs successfully conformed to a given constitution, and principle-free judgment just evaluated the better response overall without nudging a model with principles (i.e., the same judge prompt as the original MT-Bench paper). Judge prompts are described in Appendices B.2 and B.3, and example outputs are available in Appendix D.

\textbf{Results and Analysis. } \textit{Principle Alignment: }Our experiments reveal that SAMI outperforms both SFT and DPO in aligning model outputs with predefined principles. We find that SAMI wins against the SFT model 58.15\% of the time, DPO wins against the SFT model 56.50\% of the time, and SAMI wins against DPO also 57.50\% of the time. This hierarchy (SAMI > DPO > SFT) is an interesting finding. However, there is significant variance between result categories. Namely, when comparing the SAMI model with the SFT model, we find that categories math and roleplay yielded the greatest improvements, and categories coding and STEM perform the poorest.
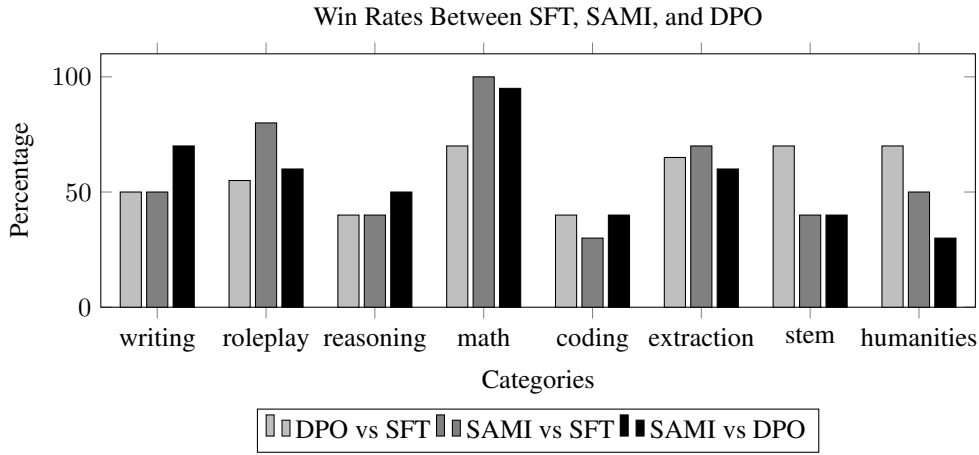
\begin{figure}[H]
    \centering
    \begin{tikzpicture}
        \begin{axis}[
            ybar,
            bar width=8pt, % Made bars thinner
            width=0.8\textwidth, % Reduced overall width
            height=0.3\textwidth, % Reduced overall height
            symbolic x coords={writing, roleplay, reasoning, math, coding, extraction, stem, humanities},
            xtick=data,
            ymin=0, ymax=110,
            ylabel={Percentage},
            xlabel={Categories},
            legend style={at={(0.5,-0.4)}, anchor=north, legend columns=-1},
            enlarge x limits={abs=0.75cm},
            title={Win Rates Between SFT, SAMI, and DPO}
        ]
        
        % DPO vs SFT (light gray)
        \addplot[fill=lightgray,draw=black] coordinates {
            (writing, 50) (roleplay, 55) (reasoning, 40) (math, 70) 
            (coding, 40) (extraction, 65) (stem, 70) (humanities, 70)
        };
        
        % SAMI vs SFT (gray)
        \addplot[fill=gray,draw=black] coordinates {
            (writing, 50) (roleplay, 80) (reasoning, 40) (math, 100) 
            (coding, 30) (extraction, 70) (stem, 40) (humanities, 50)
        };
        
        % SAMI vs DPO (black)
        \addplot[fill=black,draw=black] coordinates {
            (writing, 70) (roleplay, 60) (reasoning, 50) (math, 95) 
            (coding, 40) (extraction, 60) (stem, 40) (humanities, 30)
        };
        
        \legend{DPO vs SFT, SAMI vs SFT, SAMI vs DPO}
        \end{axis}
    \end{tikzpicture}
\end{figure}
\textit{Principle-Free Alignment: }When evaluated by the same judge as MT-Bench, we find that all three models had a similar win rate against one another, hovering between 49\% and 51\%. This result is to be expected: we bootstrap from contrastive pairs, attempting to magnify existing biases toward certain types of outputs. When the model is not pushed in either direction, it does not meaningfully affect the output quality.

\subsection{Experiment 2: Math Accuracy}

\textbf{Overview. }We note the significant improvements in the math category's ``think step by step'' principle in the previous experiment, and explore whether SAMI's stylistic improvements result in tangible improvements in accuracy. Specifically, our goal is to improve the performance of Mistral-7B on GSM8K, a large dataset of grade-school level math problems. Algorithms like STaR and rStar utilize a verifier/evaluator to only train on correct outputs and chains of thought \cite{zelikman_star_2022,qi_mutual_2024}. Here, we investigate improvement under the constraint of not verifying solutions for accuracy and use SFT as a baseline. We compare the accuracy of SFT vs. SAMI vs. SFT+SAMI in single-attempt (temperature=$0.0$) and $n$-attempt settings ($n \in \{ 10, 32 \}$; temperature=$1.0$).

\textbf{Principles and Data Generation. }We generate 4,000 prompt-completion pairs using the \verb|lighteval/MATH| dataset and principles from Appendix A on \texttt{mistral-7b} at a temperature of 0.4. To maximize contrastive pairs in the dataset, certain filters were applied. The “long response” ($LR$) from the “think step by step” principle had to be at least three times longer than the “short response” ($SR$) from the “answer concisely” principle, with $SR$ having a minimum length of 50 characters. Furthermore, the first sentence of $LR$ and $SR$ needed to differ significantly, with a Levenshtein distance of at least 0.5 times the length of $SR$. Both responses were required to terminate without looping, marked by the \texttt{} token.

\textbf{Training. }We train \texttt{mistral-7b} on the data generated SAMI for one iteration, and we also SFT on the positive samples to compare performance against a different training method. All training hyperparameters and training runs are available in Appendix C. 

\textbf{Results and Analysis. }We find that training \texttt{mistral-7b} with SAMI improves accuracy on GSM8K by 1.2\% when given 32 attempts, 3.9\% when given 10 attempts and 1.1\% for 1 attempt. SFT was more effective; SFT improved 32-attempt accuracy by 5\%, 10-attempt accuracy by 10.1\% and 1-attempt accuracy by 3.2\%. When SAMI was performed on top of SFT, 10-attempt accuracy improved by 1.3\%, 1-attempt accuracy did not change, and 32-attempt accuracy also did not change.

SAMI modestly enhances math reasoning accuracy by a few percentage points, but is less effective than simply doing SFT. Performing SAMI on top of SFT seems to have mixed results, with modest improvements for 10-attempt and 32-attempt accuracy but a slight degradation in single-attempt accuracy. Here, we hypothesize why SAMI may be ineffective for reasoning. SAMI optimizes a lower bound on the conditional mutual information  $I(y; c \mid x)$  by minimizing the row- and column-wise cross-entropy loss between the normalized log probabilities and an identity matrix. In our setup, SAMI was configured with two principles: “Write in LaTeX” and “Think step by step.” However, for mathematical reasoning tasks like those in GSM8K, the “Think step by step” principle is significantly more critical than “Write in LaTeX.” By treating both principles equally in the cross-entropy loss minimization, SAMI may dilute its focus. This misalignment means that the model does not prioritize the more important reasoning steps.
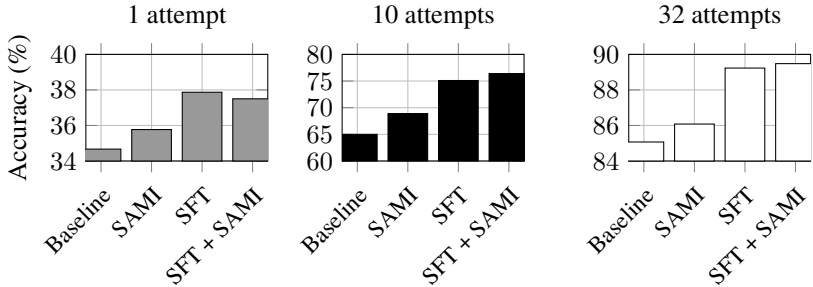
\begin{figure}[H]
    \centering
    % First bar plot for 1 attempt
    \begin{minipage}{0.2\textwidth}
        \centering
        \begin{tikzpicture}
            \begin{axis}[
                ybar,
                bar width=15pt,
                xlabel={},
                ylabel={Accuracy (\%)},
                xtick={1,2,3,4},
                xticklabels={Baseline, SAMI, SFT, SFT + SAMI},
                ymin=34, ymax=40,
                grid=major,
                width=4cm,
                height=3cm,
                x tick label style={rotate=45, anchor=north east, font=\small}, % Rotate and adjust font size
                title={1 attempt}
            ]
            % Dark green bars for 1 attempt data
            \addplot[
                ybar,
                fill=white!60!black % Darker green color
            ] coordinates {
                (1, 34.67) % Baseline
                (2, 35.77) % SAMI
                (3, 37.87) % SFT
                (4, 37.5) % SFT + SAMI
            };
            \end{axis}
        \end{tikzpicture}
    \end{minipage}
    \hspace{0.02\textwidth} % Space between the plots
    % Second bar plot for 10 attempt
    \begin{minipage}{0.2\textwidth}
        \centering
        \begin{tikzpicture}
            \begin{axis}[
                ybar,
                bar width=15pt,
                xlabel={},
                ylabel={},
                xtick={1,2,3,4},
                xticklabels={Baseline, SAMI, SFT, SFT + SAMI},
                ymin=60, ymax=80,
                grid=major,
                width=4cm,
                height=3cm,
                x tick label style={rotate=45, anchor=north east, font=\small}, % Rotate and adjust font size
                title={10 attempts}
            ]
            % Orange bars for 10 attempt data
            \addplot[
                ybar,
                fill=black
            ] coordinates {
                (1, 65)    % Baseline
                (2, 68.9)  % SAMI
                (3, 75.1)  % SFT
                (4, 76.4)  % SFT + SAMI
            };
            \end{axis}
        \end{tikzpicture}
    \end{minipage}
    \hspace{0.02\textwidth} % Space between the plots
    % Third bar plot for 32-shot (dummy data)
    \begin{minipage}{0.2\textwidth}
        \centering
        \begin{tikzpicture}
            \begin{axis}[
                ybar,
                bar width=15pt,
                xlabel={},
                ylabel={},
                xtick={1,2,3,4},
                xticklabels={Baseline, SAMI, SFT, SFT + SAMI},
                ymin=84, ymax=90,
                grid=major,
                width=4cm,
                height=3cm,
                x tick label style={rotate=45, anchor=north east, font=\small}, % Rotate and adjust font size
                title={32 attempts}
            ]
            % Blue bars for 32-shot dummy data (all 70%)
            \addplot[
                ybar,
                fill=white
            ] coordinates {
                (1, 85.07)    % Baseline
                (2, 86.08)    % SAMI
                (3, 89.23)    % SFT
                (4, 89.48)    % SFT + SAMI
            };
            \end{axis}
        \end{tikzpicture}
    \end{minipage}
\end{figure}

\section{Limitations \& Conclusion}

\textbf{Limitations. }We note that the SAMI method is most effective when the data-generating model provides enough signal to bootstrap from; when the initial model is not strong enough, there is not enough initial signal to train on. Data quality is the most significant challenge in experiment 1, and we resort to using a stronger model (\texttt{llama3.1-8b-instruct}) to generate the initial training data. We also find that SAMI is able to slightly improve math reasoning accuracy by a few percentage points, but this is far less effective than doing a simple SFT on positive samples.

\textbf{Conclusions. }We find that SAMI-generated responses are preferred to SFT-generated responses 58\% of the time, with the best results in categories including math, roleplay, and extraction. SAMI-generated responses win against DPO 57\% of the time. We note that the present results are preliminary and larger-scale validation is needed.

\textbf{Appendix. }Appendices are available online at \href{https://github.com/SohamGovande/sami-extension/blob/master/appendix.md}{this URL}.

\textbf{Acknowledgements. }We would like to thank Jan-Philipp Fränken, as well as the Computation \& Cognition Lab at Stanford University, for making this work possible.

\bibliographystyle{unsrt}  
\bibliography{references} 

\end{document}